# Traffic Flow Prediction via Variational Bayesian Inference-based Encoder-Decoder Framework

Jianlei Kong, Xiaomeng Fan, Xue-Bo Jin, and Min Zuo

*Abstract*—Accurate traffic flow prediction, a hotspot for intelligent transportation research, is the prerequisite for mastering traffic and making travel plans. The speed of traffic flow can be affected by roads condition, weather, holidays, etc. Furthermore, the sensors to catch the information about traffic flow will be interfered with by environmental factors such as illumination, collection time, occlusion, etc. Therefore, the traffic flow in the practical transportation system is complicated, uncertain, and challenging to predict accurately. This paper proposes a deep encoder-decoder prediction framework based on variational Bayesian inference. A Bayesian neural network is constructed by combining variational inference with gated recurrent units (GRU) and used as the deep neural network unit of the encoder-decoder framework to mine the intrinsic dynamics of traffic flow. Then, the variational inference is introduced into the multi-head attention mechanism to avoid noise-induced deterioration of prediction accuracy. The proposed model achieves superior prediction performance on the Guangzhou urban traffic flow dataset over the benchmarks, particularly when the long-term prediction.

## I. INTRODUCTION

IN transportation systems, sensors are widely deployed to monitor information such as vehicle speed and traffic flow—for example, urban traffic speed[1], passenger flow[2], and urban rail traffic data[3][4]. Traffic flow prediction (TFP) is the foundation of intelligent transportation systems(ITS) and is critical to implementing traffic control. Accurate TFP can help residents in travel planning and traffic control to reduce congestion and accidents.

The rapid growth in the number of private vehicles has caused more traffic accidents and traffic jams, significantly increasing the uncertainty and randomness of traffic conditions and making it challenging to predict traffic flow accurately. On the other hand, with the continuous development of computers, sensors, and cloud storage, the collected data about traffic flow can be stored in a complete record according to fixed intervals. Through data processing, the trends for traffic flow can be predicted, providing a basis for regulation and control of the traffic system and a reference for trip planning and decision-making[5], for example, helping travelers save money and time with better route guidance. Compared with short-term traffic prediction, long-term travel planning and management are needed, so medium, and long-term TFP is an essential and meaningful research area[6]. Government agencies can develop route planning based on long-term prediction to reduce traffic congestion and accidents.

Traffic time series data are usually noisy and highly nonlinear[7]. Deep learning techniques have recently been applied to TFP[8], especially long-term prediction. Still, deep learning prediction methods are an open issue. Since traffic sensors will introduce uncertainty and noise during the data collecting, such as abnormal driver operations, sensor failures, and weather changes, the prediction performance of traditional statistical methods will be degraded. At the same time, the models will be overfitted, leading to poor robustness during the training process. In addition, traffic data is periodic and trendy, making it difficult for classic deep neural networks to directly discover the intrinsic features of traffic flow data [9]. Further, for a classical neural network, the error will accumulate for long-term prediction, thus resulting in low prediction accuracy.

This paper proposes a deep encoder-decoder prediction framework based on variational Bayesian inference, aiming to overcome data noise and prediction error accumulation. The innovations of this paper are as follows:

Firstly, we introduce Bayesian inference into the recurrent neural networks and propose a Bayesian gated recurrent unit (BGRU) based on variational inference. The weights and biases of the neural network are changed to a Gaussian distribution with mean and variance, which solves the problems of poor generalization ability and low prediction accuracy of the model due to data volatility and noise.

Secondly, we incorporate the attention mechanism in the encoder-decoder framework. It can extract valid information in different subspaces and more accurately select the hidden states on all-time steps for long-term prediction. Meanwhile, the distributed multi-head attention mechanism can reduce the

This work was supported in part by the National Key Research and Development Program of China under Grant 2021YFD2100605; in part by the National Natural Science Foundation of China under Grant 62006008, Grant 62173007, and Grant 61903009; in part by the Beijing Natural Science Foundation under Grant 6214034. (*Corresponding author: Xuebo Jin; Min Zuo*.)

Jianlei Kong, Xiaomeng Fan. Xuebo Jin, Tingli Su, are with the School of Artificial Intelligence, Beijing Technology and Business University, Beijing 100048, China (e-mail: kongjianlei@btbu.edu.cn; 2130061021@st.btbu.edu.cn; jinxuebo@btbu.edu.cn;.

Min Zuo is with the School of E-Commerce and Logistics, Beijing Technology and Business University, Beijing 100048, China, and also with the National Engineering Laboratory for Agri-Product Quality Traceability, Beijing 100048, China (e-mail: zuomin@btbu.edu.cn).

computational cost and improve the computational efficiency and accuracy of the model.

The rest of the paper is arranged as follows: Section II summarizes an overview of related works on traffic time series modeling and prediction. Section III describes the general architecture of the model and process details. Section IV presents the experimental results and analysis. The experimental results show that the proposed model has good prediction performance compared with other baseline methods. The model's validity is improved by conducting practical demonstrations on the Guangzhou urban traffic flow dataset. Finally, we give the study's conclusions and discuss future research in Section V.

## II. RELATED WORK

### A. Machine Learning and Deep Neural Network

Machine learning has self-learning and nonlinear fitting capabilities, including support vector regression (SVR), matrix factorization (MF), Gaussian process (GP), and artificial neural networks (ANN)[10]-[13]. But due to uncertainty and noise, the modeling ability of machine learning is limited and not accurate enough for long-term time-series prediction.

In traffic time-series data prediction, ANNs cannot capture changes in data series when the data rapidly changes within a short period[14]-[16]. In recent years, neural networks have been pushed to new heights with the rise of deep learning. Deep learning methods can fit complex nonlinear data and thus have a solid ability to learn data[17], such as natural language processing(NLP)[18], image recognition[19], and medical diagnosis[20]. Recurrent neural networks (RNNs) have attracted much attention due to their flexibility in capturing nonlinear relationships for time series data. However, traditional RNNs have difficulty capturing long-term dependencies due to the gradient disappearance problem. In recent years, long-short-term memory networks (LSTM) and GRU have overcome this limitation [21]. Oliveira[22] uses MLP and LSTM to predict the traffic flow of an interstate highway in New York, U.S. Meng[23] proposed an LSTM for traffic speed prediction with a dynamic time-warping model, which performed better than traditional LSTM. Chen[24] designed a hybrid traffic flow prediction model based on LSTM and Sparse Auto-Encoder, which achieves a compression ratio of 20% for high-dimensional, large-scale traffic data, significantly reducing the computation complexity in TFP. Zheng[25] developed an attention-based Conv-LSTM module to predict the spatial and short-term traffic flow. In practice, the performance of the above deep neural networks would decrease rapidly in long-term prediction due to the limited capability of modeling the time series data.

### B. Encoder-Decoder Framework

The encoder-decoder is a sequence-to-sequence structure [26] using deep neural networks (e. g. CNN, RNN, or LSTM). The encoder-decoder network breaks through the limitation of the traditional RNN model with a fixed size of input and output sequences. It can extract the features of the input time series data[27]. However, as the length of the input time series increases, the information will cover the earlier one, which leads to the coded vector not reflecting the whole input vector, and lead to the prediction ability gradually decreases.

Therefore, to improve the performance of the encoder-decoder network, an attention mechanism is introduced [28]. This mechanism can assign different attention weights for all time steps and adaptively select the encoder hidden state, which is able to extract highly time-dependent useful features of the reference sequence. Attention mechanisms have been widely used in time series prediction in recent years. Jin[29] combined wavelet decomposition and bidirectional LSTM networks to integrate attention mechanisms to predict the temperature and humidity of a smart greenhouse. Lai[30] proposed a deep learning framework for multivariate time series prediction, which exploits the advantages of convolutional and recursive layers to discover the local dependence patterns between multidimensional input variables. Wang[31] integrated the attention mechanism into the seq-to-seq deep learning architecture for long-term traffic flow prediction.

### C. Bayesian Neural Network Modeling

Bayesian neural network(BNN)[32] is an inferential neural network with uncertainty. It uses Bayesian theory and the variational inference to introduce prior probabilities into the weights and biases of the neural network. It continuously adjusts the prior probabilities through backpropagation to extract the distribution characteristics hidden in the data and infer the data distribution to achieve the estimated prediction of the data distribution.

In recent years, Bayesian neural networks have been used in image detection[33], NLP[34], and time series prediction[35], etc. Zhan[36] used variational Bayesian neural network (VBNN) to validate the case of flood forecasting in the upper Yangtze River. Song[37] used variational methods to construct Bayesian linear layers to predict the Pacific coastline's maximum tsunami height. Liu[38] used Bayesian long-short term memory networks to implement fault warnings for automobile turbines. The Bayesian neural network can extract and process the hidden information of the time series and has a better description of the uncertainty of the prediction point, which is a fundamental guideline for the regulation and planning of the predicted system[39].

## III. METHODOLOGY

This paper proposes a Bayesian encoder-decoder multi-head attention model (BEDMA), which uses the Bayesian encoder-decoder model as the main structure and takes BayesianGRU as the basic unit. Incorporating the Bayesian attention mechanism, the model constructs a seq-to-seq framework based on the Bayesian encoder layer, Bayesian attention layer, and Bayesian decoder layer. In this section, we introduce these components in details.

### A. Bayesian neural network

GRU[40] is a neural network proposed by Cho in 2014, which is an improvement of LSTM. It merges forget gate and input gate into the update gate and merges memory units and hidden layers into the reset gate, which can better capture dependencies with long series. The forward propagation process of GRU is as follows:

$$z_t = \sigma(W_z[h_{t-1}, x_t] + b_z) \quad (1)$$
$$r_t = \sigma(W_r[h_{t-1}, x_t] + b_r) \quad (2)$$
$$\tilde{h}_t = \tanh(W_h[r_t \odot h_{t-1}, x_t] + b_h) \quad (3)$$
$$h_t = (1 - z_t) \odot \tilde{h}_t + z_t \odot h_{t-1} \quad (4)$$

where $W_r$ and $b_r$ are the weights and biases of the reset gate, which decides whether the previous hidden state $h_{t-1}$ is ignored; $W_z$ and $b_z$ are the weights and biases of the update gate, which controls how much information from the previous hidden state will carry over to the current hidden state $h_t$; $W_h$ and $b_h$ are used to calculate the current candidate hidden state.

The weights $W$ and biases $b$ of the traditional GRU are fixed values, which makes it overfitting in predicting noisy traffic flow data such that it cannot fit other data or predict future observations well. Inspired by BNN, we combine variational inference and GRU to Bayesian GRU(BGRU) so that its weights and biases become sampling points with Gaussian distribution to avoid the overfitting on noisy data.

BNN uses the probability distribution over the network weights and outputs the prediction in that distribution. BNN can provide a probability solution to the uncertainty problem in conventional networks' training process. The neural network treats the dataset as a probabilistic model $P(y|x,w)$: given the inputs $x$ and weights $w$, the neural network has a probability for each output $y$. And the BNN computes the posterior distribution of weights given the training data $P(w|D)$, which is derived from

$$P(y|x) = \mathrm{E}_{P(w|D)}[P(y|x,w)] \quad (5)$$

where $D = \{(x_1, y_1), (x_2, y_2), \cdots, (x_m, y_m)\}$ is the dataset. An infinite number of weights are obtained based on the posterior distribution, all of which predict the unknown label $y$ for a given test data item $x$.

Therefore, taking an expected value under the posterior distribution of weights is equivalent to using an infinite set of neural networks. Variational learning finds the parameters of the distribution $\theta \sim \mathcal{N}(\mu, \sigma)$ over the weights $q(w|\theta)$ such that the Kullback-Leibler divergence(KL) of the weights concerning the Bayesian posterior probability is minimized as follows:

$$\begin{aligned}\theta^* &= \arg\min_\theta KL[q(w|\theta) \| P(w|D)] \\ &= \arg\min_\theta \int q(w|\theta) \log[q(w|\theta)/P(w|D)]dw \\ &= \arg\min_\theta \int q(w|\theta) \log[q(w|\theta)/P(D|w)P(w)]dw \\ &= \arg\min_\theta KL[q(w|\theta) \| P(w)] - \mathrm{E}_{q(w|\theta)}[\log P(D|w)]\end{aligned} \quad (6)$$

where $P(D)$ can be ignored because it is not related to $\theta$.

Assuming that the parameters are independent of each other, the loss function of the network which can be approximated using Monte Carlo sampling is as follows:

$$F_{loss}(D, \theta) \approx \frac{1}{n}\sum_{i=1}^{n} \log q(w^{(i)}|\theta) - \log P(w^{(i)}) - \log P(D|w^{(i)}) \quad (7)$$

The BGRU has the same chain structure as the GRU, with the difference that each weight $W$ and bias $b$ in the BGRU is a distribution with trainable parameters. The initialized distribution of them is a standard normal distribution, and its optimal mean and variance are obtained by training. When using BGRU, predictions are obtained by sampling over the distribution of weights. The mean of the predictions is used as the final prediction, and the variance of the multiple predictions is used as the confidence interval. The structure of the BGRU is shown in Fig. 1.

Taking $W_r$ and $b_r$ as an example. Let $W_n^{(i)}$ denote the $n$-th sampling weight of the $i$-th layer and $b_n^{(i)}$ denote the bias, which conforms to the normal distribution. To ensure that the variance is non-negative and the standard deviation is derivable, a Gaussian distribution with mean $\mu^{(i)}$ and standard deviation $\sigma^{(i)}$ is used for the translation and scaling transformation. The final goal is to optimize $\mu^{(i)}$ and $\sigma^{(i)}$.

$$W_{(n)}^{(i)} = \mathcal{N}(0,1) * \log(1 + \sigma^{(i)}) + \mu^{(i)} \quad (8)$$
$$b_{(n)}^{(i)} = \mathcal{N}(0,1) * \log(1 + \sigma^{(i)}) + \mu^{(i)} \quad (9)$$

We use BGRU as the encoder for the codec structure, the encoder network consists of multiple layers of BGRUs, and its forward propagation process is as follows:

$$\begin{aligned}w_z &= \mathcal{N}(0,1) * \log(1 + \sigma_{zw}) + \mu_{zw} \\ b_z &= \mathcal{N}(0,1) * \log(1 + \sigma_{zb}) + \mu_{zb} \\ z_t &= \sigma(w_z[h_{t-1}, X_{input}^i] + b_z)\end{aligned} \quad (10)$$

$$\begin{aligned}w_r &= \mathcal{N}(0,1) * \log(1 + \sigma_{rw}) + \mu_{rw} \\ b_r &= \mathcal{N}(0,1) * \log(1 + \sigma_{rb}) + \mu_{rb} \\ r_t &= \sigma(w_r[h_{t-1}, X_{input}^i] + b_r)\end{aligned} \quad (11)$$

$$\begin{aligned}w_h &= \mathcal{N}(0,1) * \log(1 + \sigma_{hw}) + \mu_{hw} \\ b_h &= \mathcal{N}(0,1) * \log(1 + \sigma_{hb}) + \mu_{hb} \\ \tilde{h}_t &= \tanh(w_h[r_t \odot h_{t-1}, X_{input}^i] + b_h)\end{aligned} \quad (12)$$

$$h_t = (1 - z_t) \odot \tilde{h}_t + z_t \odot h_{t-1} \quad (13)$$

The loss function is defined as follows:

$$loss = \log(q(w|\theta)) - \log(p(w)) \quad (14)$$

where $p(w)$ is a custom a priori distribution and $\theta \sim \mathcal{N}(u, \sigma^2)$ and $q(w|\theta)$ are posterior distributions. This allows the BGRU to learn the distribution features, while the target given during training is a sequence of deterministic values so that deterministic errors need to be incorporated.

$$loss = \alpha \bullet mse(\hat{y}, y) + \frac{1}{\alpha} \bullet [\log(q(w|\theta)) - \log(p(w))] \quad (15)$$

where $\hat{y}$ is the prediction of the output under the current weight sampling, and $\alpha$ is the weight coefficient, which is equal to the product of the number of training samples and the batch size.

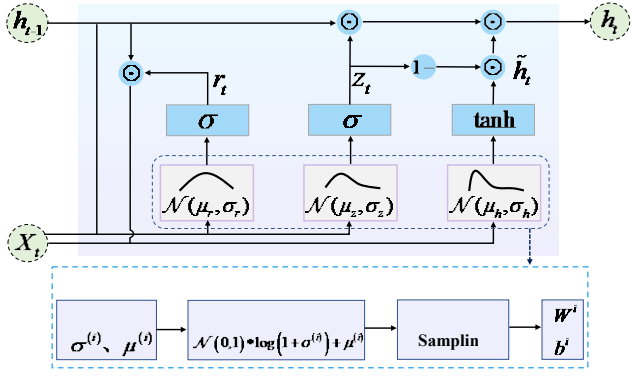

**Fig. 1.** BGRU structure, where $W_r$ and $b_r$ are the sampling point of a Gaussian distribution obeying the mean $\mu_z$ and covariance $\sigma_z$.

*B. Bayesian Multi-head Attention*

The encoder-decoder model can extract the input data's information and is widely used in NLP. But the coding information will be lost in a long data series, resulting in the coding vector not reflecting the information of the whole input, thus reducing the prediction accuracy.

The attention mechanism is a modification to the encoder-decoder model. As compared to the base encoder-decoder model, the output of the attention mechanism in the encoder is involved in the computation of each step in the decoder. The encoder-decoder model with attention mechanism has removed the bottleneck of fixed-length coding, and the loss information from the encoder to the decoder will be low. The structure of the encoder-decoder model based on the attention mechanism is shown in Fig. 2.

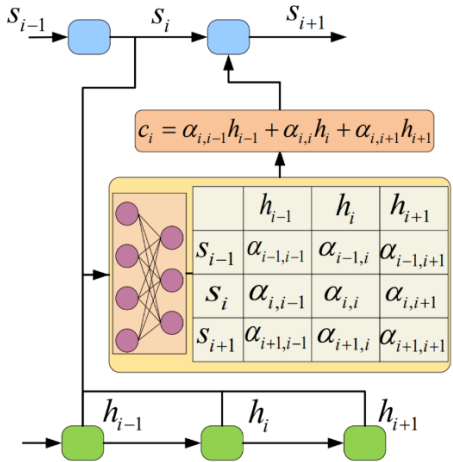

**Fig. 2** Addictive attention mechanism model

The encoding vector $c_i$ depends on a sequence of annotations $h = [h_1, h_2, \cdots, h_j, \cdots, h_t]^T \in \mathbb{R}^{t*m}$ to which the encoder maps the input statement $X = [X_1, X_2, \ldots, X_t]^T \in \mathbb{R}^{t*n}$. The context vector $c_i$ is computed as a weighted sum of these annotations $h_j$.

$$c_i = \sum_{j=1}^{t} \alpha_{ij} h_j \quad (16)$$

The weight of each annotation $h_j$, $\alpha_{ij}$ is calculated as follows:

$$\alpha_{ij} = \frac{\exp(e_{ij})}{\sum_{k=1}^{t} \exp(e_{ik})} \quad (17)$$

where $e_{ij}$ is used to calculate the match between the decoder's hidden state $s_{i-1}$ and the encoder's $j$ hidden state $h_j$, which is calculated as follows:

$$e_{ij} = v_e^\top \tanh(W_e s_{i-1} + U_e h_j) \quad (18)$$

where $v_e \in R^t, W_e \in R^{t \times m}, U_e \in R^{t \times t}$ is the parameter to be learned. The softmax function is applied $e_{ij}$ to ensure that all attention weights sum to 1.

Fig. 3. show the structure of the Bayesian multi-head attention mechanism by transforming the linear layer of multi-head attention into a Bayesian linear layer.

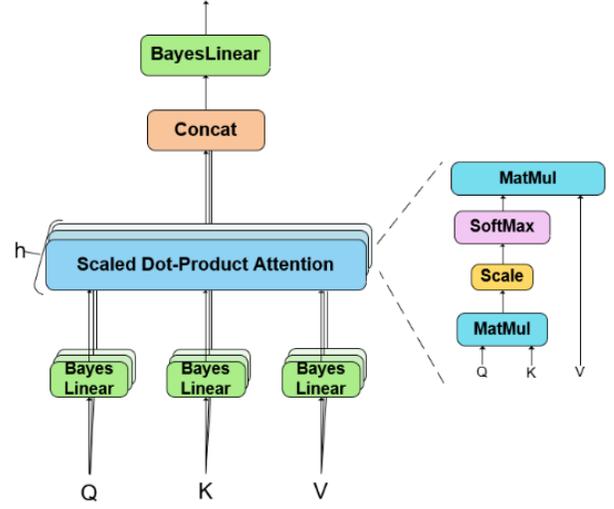

**Fig. 3** Structure of Bayesian multi-head attention model

In the multi-head attention mechanism, multi-heads refer to multiple scaled dot-product calculations of the input Query($Q$), Key($K$) and Value($V$). The weights of different heads are independent of each other and are not shared.

Define $d = m/h$, $q = k = v = H$, and $h$ is the number of heads. The encoder output $H$ is transformed $h$ times, and then the input of the $i$-the head is $Q_i, K_i, V_i \in \mathbb{R}^{t \times d}$.

$$Q_i = w_{iq} q$$
$$w_{iq} = \mathcal{N}(0,1) * \log(1 + \rho_i) + \mu_i \quad (19)$$

$$K_i = w_{ik} k$$
$$w_{ik} = \mathcal{N}(0,1) * \log(1 + \rho_k) + \mu_k \quad (20)$$

$$V_i = w_{iv} v$$
$$w_{iv} = \mathcal{N}(0,1) * \log(1 + \rho_v) + \mu_v \quad (21)$$

where $\mu_i, \rho_i, \mu_k, \rho_k, \mu_v, \rho_v$ is the parameter to be learned.

Attention mechanism is calculated in three steps: firstly, the similarity between the $Q$ vector and each $K$ is calculated, and the corresponding weights are obtained; then, these weights are normalized using the softmax function; finally, the weights and the corresponding $V$ are weighted and summed to obtain the attention weights. The attention weight calculation process and the output of the $i$-th $head_i \in \mathbb{R}^{t \times d}$ are as follows:

$$AttentionWight = \text{soft} \max\left(\frac{Q \bullet K^T}{\sqrt{d}}\right) \quad (22)$$

$$head_i = \text{soft}\max\left(\frac{Q_i \bullet K_i^T}{\sqrt{d}}\right) \bullet V_i \quad (23)$$

where $\bullet$ means dot-product, $head_i$ is the weighted Bayesian encoder output.

In multi-head attention, the $Q$, $K$ and $V$ undergo a linear calculation, and then are input to the scaled dot-product attention, which requires h dot product scaling. The so-called multi-head is to perform multiple dot product scaling, each dot product scaling is a head, the parameters are not shared between different heads, and $Q$, $K$ and $V$ are different for each linear transformation. After passing the h dot product scaling process, the attention result is concatenated, and the encoding vector $C$ obtained after linear transformation is the result of multi-head attention.

It can be seen that multi-head attention performs the dot-product multiple times instead of once, and this multiple dot-product process can learn from different subspaces. Multi-head attention uses multiple queries $Q$ to extract multiple information from the input data $X_{input}^i$ by parallel computation. Each $head_i$ focuses on a different part of the input information. The distributed multi-head attention mechanism can save computational resources, reduce computational costs, and improve the computational efficiency of the model.

*C. Model for Traffic Flow Prediction*

The model framework is shown in Fig. 4, in which the encoder-decoder is constructed using BGRU as the basic neural network, and the multi-head attention mechanism based on variational inference is fused in the encoder-decoder.

Firstly, a sliding window is applied to the data to obtain the network input sequence $X_{input}^i = [x_{s*(i-1)+1},\cdots,x_{s*(i-1)+t}]$, which $t$ is the length of the input sequence. Traffic flow data is transmitted to an encoder consisting of a BGRU. During propagation, the hidden states of the BNN layer are output at each time step to obtain the encoder output $H = [h_1, h_2, \cdots, h_t]^T \in \mathbb{R}^{t \times m}$.

After getting the Bayesian encoder output, the Bayesian encoder is transmitted to the Bayesian attention layer. Through the Bayesian attention layer, feature information of traffic flow data is obtained and different attention is paid to feature information, which further enhances the information extraction capability. Then output of heads are concatenated together and subjected to Bayesian linear variation to obtain the encoding vector $C \in \mathbb{R}^{t \times m}$.

$$\begin{aligned} C &= w_c \bullet [head_0; head_1; \ldots; head_q] \\ w_c &= \mathcal{N}(0,1) * \log(1+\rho_c) + \mu_c \end{aligned} \quad (24)$$

where $\mu_c, \rho_c$ is the parameter to be learned.

The Bayesian decoder also consists of multiple layers of BGRUs. The coding vector $C$ is input to the Bayesian decoder, and after passing through the layers, the hidden state of the last time step in the Bayesian decoder $h' \in \mathbb{R}^m$ is output, and a nonlinear transformation is performed to obtain the prediction sequence.

$$\begin{aligned} \tilde{Y} &= [\tilde{y}_1, \tilde{y}_2, \cdots, \tilde{y}_\tau]^T = relu(w_y h' + b_y) \\ w_y &= \mathcal{N}(0,1) * \log(1+\rho_y) + \mu_y \end{aligned} \quad (25)$$

where $\mu_y, \rho_y$ is the parameter to be learned and $r$ is the predicted length of the target sequence. $relu$ is the activation function with the following expression.

The prediction sequence $\tilde{Y} = [\tilde{y}_1, \tilde{y}_2, \cdots, \tilde{y}_\tau]^T$ is compared with the target sequence $X_{target}^i = [x_{s*(i-1)+t+1},\cdots,x_{s*(i-1)+t+\tau}]$ to calculate the error, and the error back propagation is used to update the model parameters to minimize the error between the predicted sequence and the target sequence.

The model is trained on the network using the training dataset and then evaluated using the validation dataset to minimize overfitting. All models are optimized using the adaptive moment estimation (Adam) optimization algorithm, which uses momentum and adaptive learning rates to accelerate convergence with high computational efficiency and a small memory footprint.

The following algorithm 1 gives the training process of the model:

| Algorithm 1 Training Algorithm of BEDMA Model |
|---|
| **Input:** the traffic flow data $X_{input}^i$, *window_size*, *epochs*, *batch_size* |
| **Output:** the prediction of traffic flow $\tilde{Y} = [\tilde{y}_1, \tilde{y}_2, \cdots, \tilde{y}_\tau]^T$ |
| 1: Data preprocessing |
| 2: **for** *i*=0 to *epochs* **do** |
| 3:     Calculate $H = [h_1, h_2, \cdots, h_t]^T \in \mathbb{R}^{t \times m}$ by Eq. (10)-(13), draw hyperparameters $\{W, b, \mu$ and $\sigma\}$. |
| 4:     Calculate encoding vector $C \in \mathbb{R}^{t \times m}$ by Eq. (19)-(24), draw hyperparameters $\{\mu_i, \rho_i, \mu_k, \rho_k, \mu_v, \rho_v, \mu_c, \rho_c\}$. |
| 5:     Calculate the predicted traffic flow by Eq. (25). |
| 6:     Optimizing loss by MAE and Adam. |
| 7:     **if** reached the criterion then **break** |

IV. EXPERIMENTAL AND RESULT ANALYSIS

*A. Dataset*

This experiment uses the Guangzhou urban traffic flow dataset, which consists of traffic speeds in urban highway and main road, from August 1, 2016, to September 30, 2016, with a data deficiency rate of 1.29%. During the experiment, traffic flow data from the first 48 days were used as training samples for the model, and data from the next 13 days as testing samples. Part of the data visualization is shown in Fig. 5.

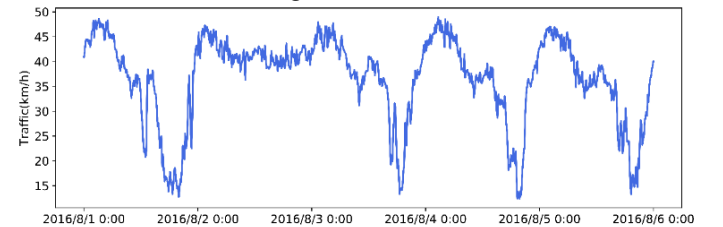

**Fig. 5** Data visualization

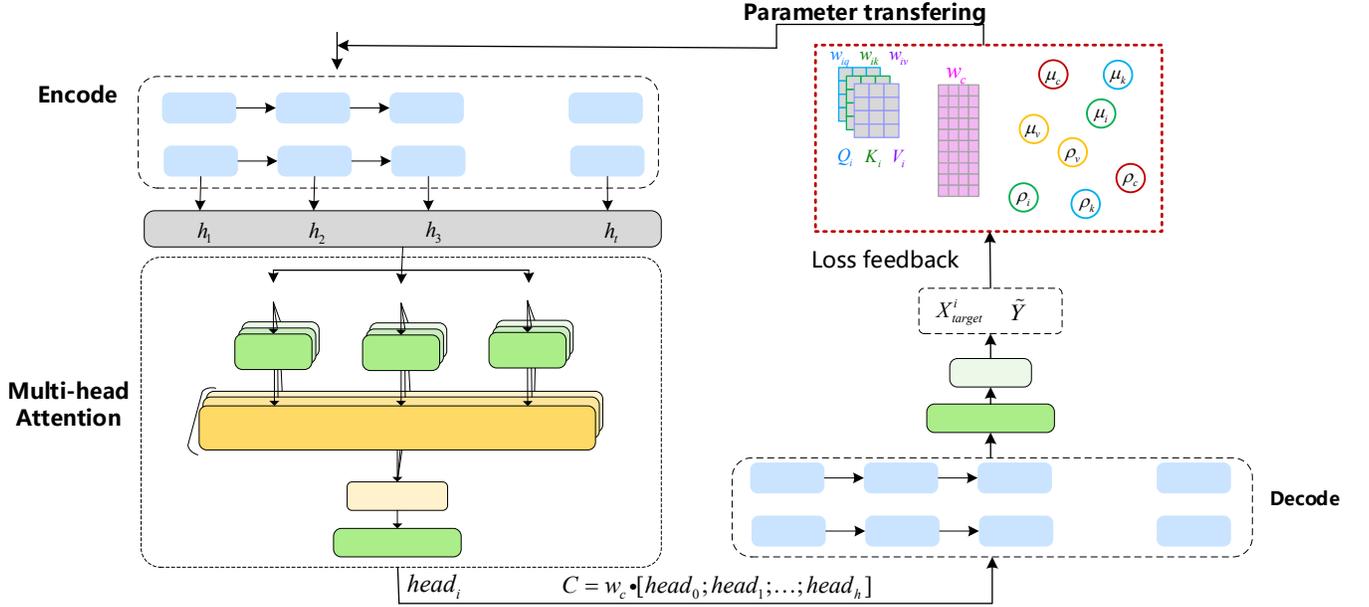

**Fig. 4** Bayesian encoder-decoder multi-head attention model framework

## B. Evaluation Metrics

The experiments used root mean squared error (RMSE) Pearson's correlation coefficient R, and symmetric mean absolute percentage error (SMAPE) as the indexes and indicators for evaluating the model. RMSE calculates the sum of squares of the distance between the prediction and actual traffic speed, calculates the deviation between them, and reflects the degree of dispersion of the prediction. The smaller the value of RMSE, the smaller the deviation is. R measures the correlation between the prediction and actual values. R has a maximum of 1, and the closer it is to 1, the better the regression line fits the actual values. SMAPE is used to measure the proportion of deviation between the prediction and actual traffic speed. The formulas of these three indicators are as the following.

$$RMSE = \sqrt{\frac{1}{m}\sum_{i=1}^{m}(y_i - \hat{y}_i)^2} \quad (25)$$

$$SMAPE = \frac{100\%}{m}\sum_{i=1}^{m}\frac{|\hat{y}_i - y_i|}{(\hat{y}_i + |y_i|)/2} \quad (26)$$

$$R = \frac{\sum_{t=1}^{T}(\hat{y}_t - \bar{\hat{y}}_t)(y_t - \bar{y}_t)}{\sqrt{\sum_{t=1}^{T}(\hat{y}_t - \bar{\hat{y}}_t)^2 \sum_{t=1}^{T}(y_t - \bar{y}_t)^2}} \quad (27)$$

where $y_i$ is the actual traffic speed, $\hat{y}_i$ is the prediction, $m$ represents the number of samples, $\bar{y}_i$ is the average of the actual traffic speed, and $\bar{\hat{y}}_i$ is the average of the prediction.

## C. Comparison and Analysis

In order to verify the performance of the model, we compare the proposed model with Linear[41], RNN[42], GRU[43], LSTM[44], En-Decoder[45], Attention[46], and Multi-head attention (Mhatt)[47]. The training parameters of the model are set as follows: the number of iterations is 100, the optimizer is Adam, and the learning rate is 0.001. The number of layers is 2, the number of units per layer is 64. Batch size set to 12, the attention model encoder and decoder are 1 each, the number of attention heads is 2.

All models were written in Python 3.8 environment based on Pytorch deep learning framework. All experiments were done on a server with the following parameters: Ubuntu 20.04 bit-64 operating system; Intel Core i7-6800 processor CPU @3.4GH; NVIDIA GTX1080Ti 16G. To verify the stability of the models, each model was repeated 10 times independently. The evaluation of model prediction performance is performed using the evaluation metrics mentioned in Section IV.*B*.

*1) Comparison of Different Prediction Intervals.* We developed a BEDMA model for medium and long-term traffic speed prediction (10min/30min/60min.) The performance is compared in Table I. From the experimental results, each index of the proposed model outperform the baseline models. The RMSE and SMAPE of the proposed model are lower than those of the other models, which indicates that the model has the smallest difference between the predicted and actual ones. The R indicators are greater than the other models, showing that the model has the highest goodness-of-fit.

Specifically, it can be seen from Fig. 6 that in the short-term (10 minutes) prediction, the linear model performs the worst, and the Mhatt model performs the best among the baseline models, with RMSE and SMAPE of 1.4772 and 0.0503, respectively. And, the proposed BEDMA model have a good performance in both short and long-term predictions. The RMSE and SMAPE are lower than the Mhatt model by 1% and 3%, respectively. To further test the model's effectiveness in long-term prediction, we conducted a long-term prediction test to predict the traffic speed for the following 30 and 60 minutes. The accuracy of each model decreases as the prediction time increases, and then BEDMA still outperforms the other baseline models and shows better stability in long-term prediction.

TABLE I
EVALUATION OF DIFFERENT PREDICTION MODELS

| Model | 10 minute | | | 30 minute | | | 60 minute | | |
|---|---|---|---|---|---|---|---|---|---|
| | RMSE | SMAPE | R | RMSE | SMAPE | R | RMSE | SMAPE | R |
| Linear[46] | 1.5745 | 0.0518 | 0.9689 | 2.7582 | 0.0812 | 0.9337 | 3.3732 | 0.1003 | 0.8677 |
| RNN[33] | 1.5734 | 0.0514 | 0.9691 | 2.4028 | 0.0727 | 0.9326 | 3.0949 | 0.0909 | 0.8821 |
| GRU[34] | 1.5292 | 0.0503 | 0.9704 | 2.2934 | 0.0689 | 0.9392 | 3.1159 | 0.0919 | 0.8896 |
| LSTM[35] | 1.5082 | 0.0497 | 0.9716 | 2.3102 | 0.0691 | 0.9384 | 3.0768 | 0.0905 | 0.8899 |
| En-Decoder[36] | 1.5028 | 0.0499 | 0.9714 | 2.2808 | 0.0684 | 0.9401 | 2.9617 | 0.0877 | 0.8955 |
| Attention[37] | 1.4968 | 0.0496 | 0.9714 | 2.2929 | 0.0691 | 0.9380 | 2.9453 | 0.0876 | 0.8969 |
| Mhatt[47] | 1.4946 | 0.0503 | 0.9718 | 2.2449 | 0.0680 | 0.9403 | 2.9142 | 0.0875 | 0.8969 |
| **BEDMA** | **1.4772** | **0.0486** | **0.9721** | **2.2211** | **0.0661** | **0.9415** | **2.8813** | **0.0858** | **0.9006** |

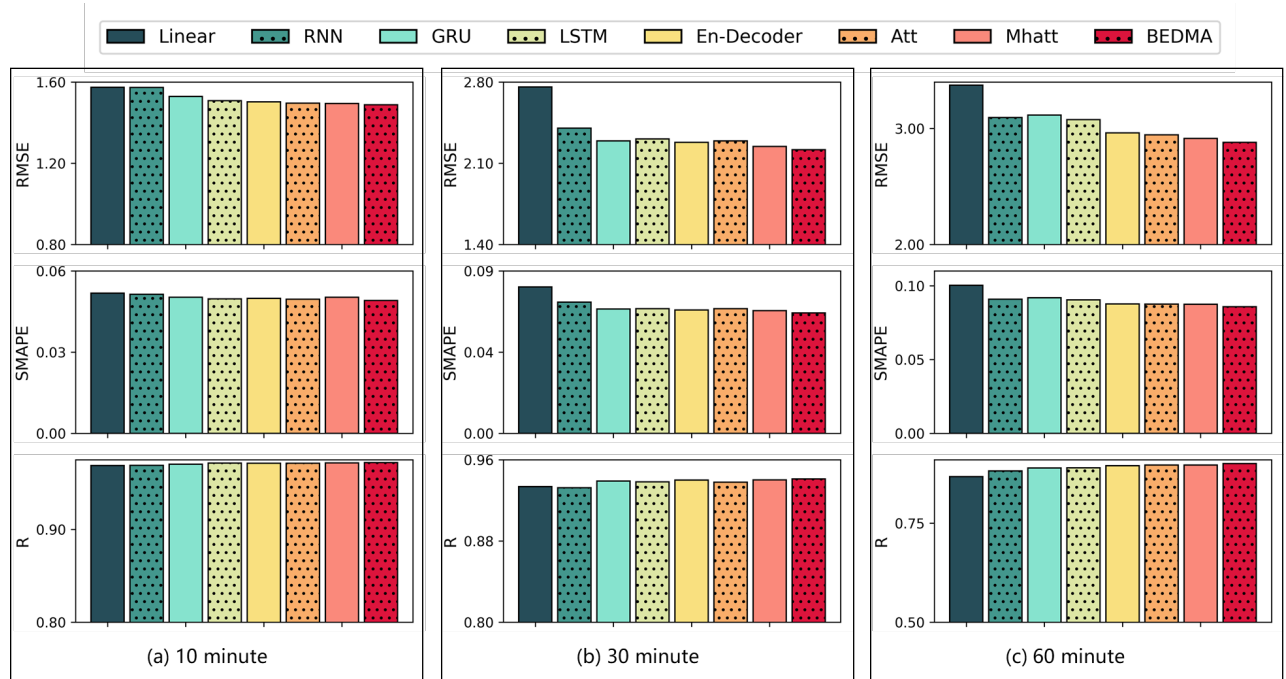

Fig. 6 Traffic speed prediction evaluation index of different intervals

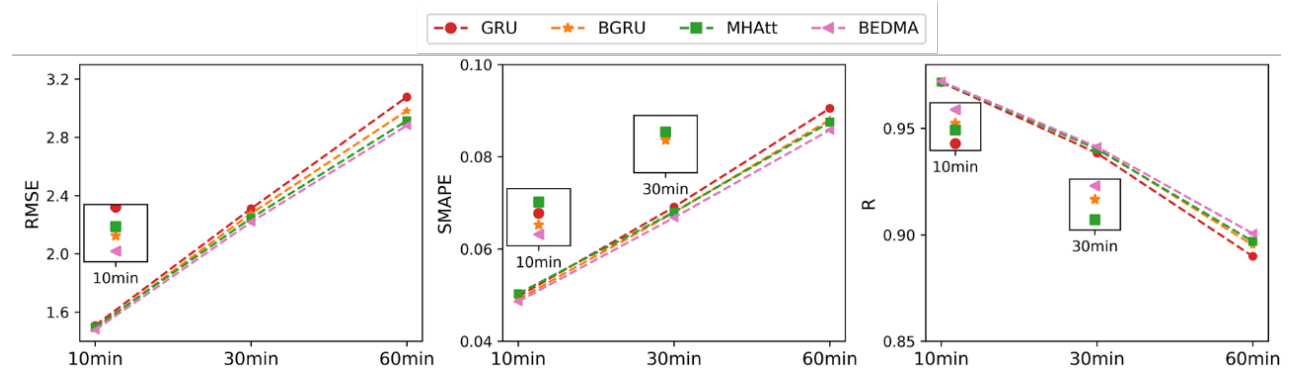

Fig. 7 Evaluation metrics of different prediction intervals for each model

*2) Ablation Study.* In order to validate the Bayesian encoder-decoder model based on the attention mechanism proposed, the modeling predictions are validated using the traffic speed dataset. The same arrangement as the comparison experiments was used to set up. We compared the GRU and BayesianGRU, Mhatt, and BEDMA models for each prediction steps (10 min/30 min/60 min).

As we can see from Table II, the RMSE of BGRU is 1.3% lower than GRU in predicting the traffic flow for the next 10 minutes compared to GRU. As the prediction steps increases, the RMSE of BGRU decreases by 3.1% compared to GRU in predicting the traffic flow for 60 minutes. The experimental results show that adding variational inference improves the model fitting ability, fits the noisy data well and reduces the error. By comparing the experiment results of Mhatt and BEDMA, the RMSE of BEDMA is reduced by more than 1.1% compared with Mhatt at different prediction times, and the R is higher than Mhatt, which indicates that the incorporation of variational inference into the attention mechanism can improve the feature extraction ability of the model. Comparing the experimental results of GRU and BEDMA, the RMSE of BEDMA is reduced by 2%-6.4% compared with GRU. By combining variational inference with GRU and attention, the model can substantially improve its ability to model noisy traffic flow data and reduce errors.

TABLE II
EXPERIMENTAL RESULTS OF ABLATION STUDY

| Model | 10 minute | | 30 minute | | 60 minute | |
|---|---|---|---|---|---|---|
| | RMSE | R | RMSE | R | RMSE | R |
| GRU | 1.5082 | 0.9716 | 2.3102 | 0.9384 | 3.0768 | 0.8899 |
| BGRU | 1.4881 | 0.9719 | 2.2759 | 0.9409 | 2.9820 | 0.8952 |
| Mhatt | 1.4946 | 0.9718 | 2.2449 | 0.9403 | 2.9142 | 0.8969 |
| BEDMA | 1.4774 | 0.9721 | 2.2176 | 0.9413 | 2.8813 | 0.9006 |

In addition, we visualize the traffic flow prediction results for three roads and analyze the model's prediction performance on different data sets. Fig. 7 shows the curves of the models for different prediction intervals as10, 30 and 60 minutes. In summary, the experimental results demonstrate that the model has excellent long-term prediction performance under prediction intervals.

*3) Performance Under Different Roads*

We selected three roads for traffic flow prediction, and Table III show the prediction results of the three roads for 10 minutes. We can see that the error indicators are all at a low level. Moreover, the R indexes all reach above 0.97, indicating that the prediction fit the actual traffic flow well.

Fig. 8-Fig. 10 show the prediction results separately, and it can be seen that the predicted and actual curves fit well. In summary, the experiments demonstrate that the model has excellent prediction performance under different roads.

TABLE III
PREDICTION PERFORMANCE FOR THREE ROADS

| Model | 10 minute | | 30 minute | | 60 minute | |
|---|---|---|---|---|---|---|
| | RMSE | R | RMSE | R | RMSE | R |
| Road 1 | 1.3833 | 0.9746 | 2.1043 | 0.9502 | 2.9234 | 0.9093 |
| Road 2 | 1.1261 | 0.9805 | 1.7683 | 0.9584 | 2.3587 | 0.9292 |
| Road 3 | 1.3344 | 0.9762 | 2.1045 | 0.9476 | 2.7638 | 0.9202 |

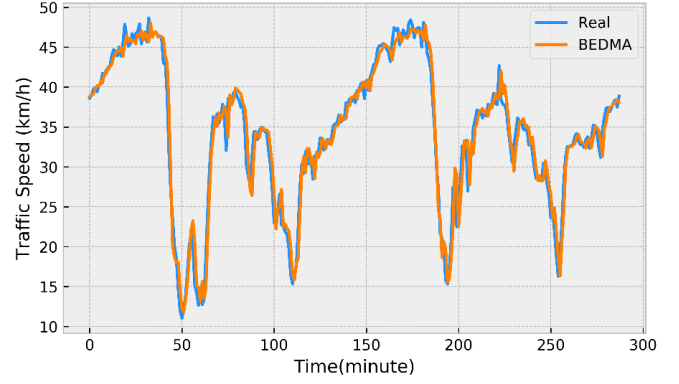

Fig. 8 Prediction result of traffic flow in Road 1 over a 10 minute

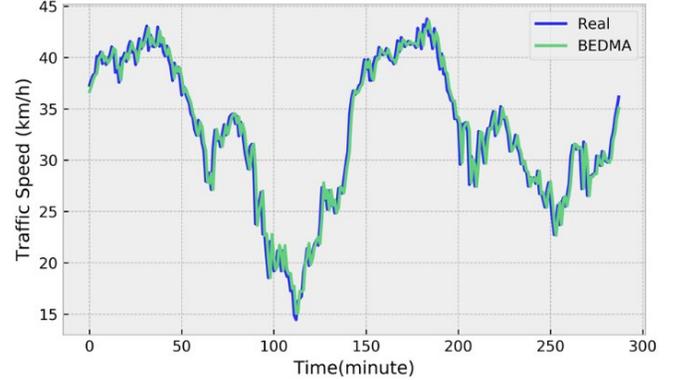

Fig. 9 Prediction result of traffic speed on Road 2 over a 10 minute

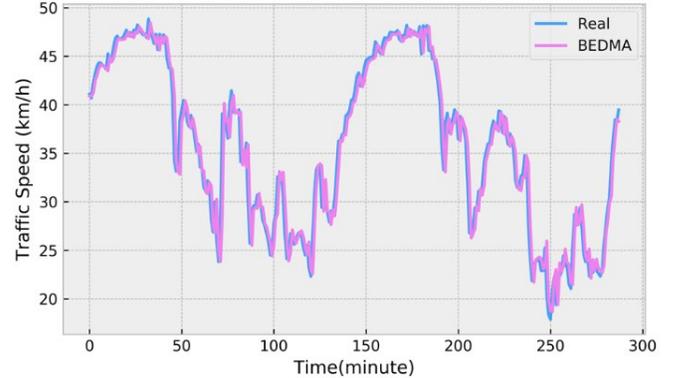

Fig. 10 Prediction result of traffic speed on Road 3 over a 10 minute

V. CONCLUSION

In this paper, an attention encoder-decoder model based on a Bayesian neural network is proposed, in which Bayesian neural network is used as the primary neural network unit within encoder and decoder framework while incorporating a multi-headed attention structure based on variational inference. The model is validated on Guangzhou urban traffic flow dataset and has better prediction performance compared to other baseline models with different prediction interval. Under different prediction interval, such as 10 minutes, 30 minutes and 60 minutes, the R evaluation index is 0.9721, 0.9415 and 0.9006, respectively, and the RMSE decreases to 1.4772, 2.2211, and 2.8813. It was further confirmed by ablation experiments that the inclusion of variational inference improved the R-evaluation metrics of the model while reducing each error evaluation metric. The framework performs better prediction for nonlinear and noisy traffic flow data than the benchmark model. It can fully exploit the internal correlation of the data and accurately predict the changing trend. In the next work, the model will be further optimized to expand the application of the model to other types of time series data, while the introduction of Bayesian inevitably increases the computational cost.